\documentclass[letterpaper, 10 pt, conference, twoside]{ieeeconf}
\IEEEoverridecommandlockouts  
\usepackage{graphicx}
\usepackage{stfloats}
\usepackage{caption}
\usepackage{amsfonts}
\usepackage{array}
\usepackage[ruled]{algorithm2e}
\usepackage[colorlinks,breaklinks=true]{hyperref}
\usepackage{cite}
\usepackage{amsmath,amsfonts,amssymb}

\newcommand{\eg}{{\em e.g.\ }}

\newcommand{\et}{{\em et al.\ }}

\newcommand{\beq}{\begin{equation}}
\newcommand{\eeq}{\end{equation}}
\newcommand{\bear}{\begin{eqnarray}}
\newcommand{\bears}{\begin{eqnarray*}}
\newcommand{\eear}{\end{eqnarray}}
\newcommand{\eears}{\end{eqnarray*}}
\newcommand{\bdm}{\begin{displaymath}}
\newcommand{\edm}{\end{displaymath}}
\newcommand{\lba}{\left[\begin{array}}
\newcommand{\ear}{\end{array}\right]}

\usepackage[normalem]{ulem}
\useunder{\uline}{\ul}{}
\usepackage{booktabs}
\usepackage{multirow}
\usepackage{etoolbox}
\makeatletter
\patchcmd{\@makecaption}
{\scshape}
{}
{}
{}
\makeatletter
\patchcmd{\@makecaption}
{\\}
{.\ }
{}
{}
\makeatother
\def\tablename{Table}
\title{\LARGE \bf
	A Graph Attention Spatio-temporal Convolutional Network for 3D Human Pose Estimation in Video}
\author{Junfa Liu$^{1\dag}$, Juan Rojas$^{2\dag\ast}$, Zhijun Liang$^{1}$, Yihui Li$^{1}$, and Yisheng Guan$^{1\ast}$ 
	\thanks{$^{1}$The Biomimetic and Intelligent Robotics Lab (BIRL), School of Electromechanical Engineering, Guangdong University of Technology, 510006 Guangzhou, China. $^{2}$Dept. of Mechanical and Automation Engineering,  Chinese University of Hong Kong, Hong Kong, China. $\dag$ Equal contribution. $\ast$ Corresponding authors (\url{ysguan@gdut.edu.cn} and \url{juan.rojas@cuhk.edu.cn}). The work in this paper is in part supported by the  Frontier and Key Technology Innovation Special Funds of Guangdong Province (Grant No. 2017B050506008) and the Key R\&D Program of Guangdong Province (Grant No. 2019B090915001).}
	}

\begin{document}
\maketitle
\thispagestyle{empty}
\pagestyle{empty}
\begin{abstract}
Spatio-temporal information is key to resolve occlusion and depth ambiguity in 3D pose estimation. Previous methods have focused on either temporal contexts or local-to-global architectures that embed fixed-length spatio-temporal information. To date, there have not been effective proposals to simultaneously and flexibly capture varying spatio-temporal sequences and effectively achieves real-time 3D pose estimation.
In this work, we improve the learning of kinematic constraints in the human skeleton: posture, local kinematic connections, and symmetry by modeling local and global spatial information via attention mechanisms. To adapt to single- and multi-frame estimation, the dilated temporal model is employed to process varying skeleton sequences. Also, importantly, we carefully design the interleaving of spatial semantics with temporal dependencies to achieve a synergistic effect. 
To this end, we propose a simple yet effective graph attention spatio-temporal convolutional network (GAST-Net) that comprises of interleaved temporal convolutional and graph attention blocks.
Experiments on two challenging benchmark datasets (Human3.6M and HumanEva-I) and YouTube videos demonstrate that our approach effectively mitigates depth ambiguity and self-occlusion, generalizes to half upper body estimation, and achieves competitive performance on 2D-to-3D video pose estimation. Code, video, and supplementary information is available at: \href{http://www.juanrojas.net/gast/}{http://www.juanrojas.net/gast/}
\end{abstract}


\section{INTRODUCTION}
3D human pose estimation from video is a very active research area that impacts domains like action recognition, virtual reality, and human-robot interaction. Previously, 3D pose estimation was computed using depth sensors, motion capture, or multi-view images in indoor environments. However, with recent advances in 2D human pose estimation through deep learning along with massive availability of in-the-wild data, there has been great progress in solving 3D pose estimation from monocular images \cite{zhao2019semantic}. Recent works \cite{martinez2017simple,dabral2018learning,wandt2019repnet} only use 2D human pose representations and achieve competitive performance and generalization, which avoid the influence of background noise and human external appearance. Additionally, compared to processing RGB images, 2D-to-3D methods leverage 2D keypoints (which use less computation) and enable longer-term frame estimations. In this paper, we focus on 2D-to-3D estimation. 

\begin{figure}[t]
	\centering
	\includegraphics[scale=0.11]{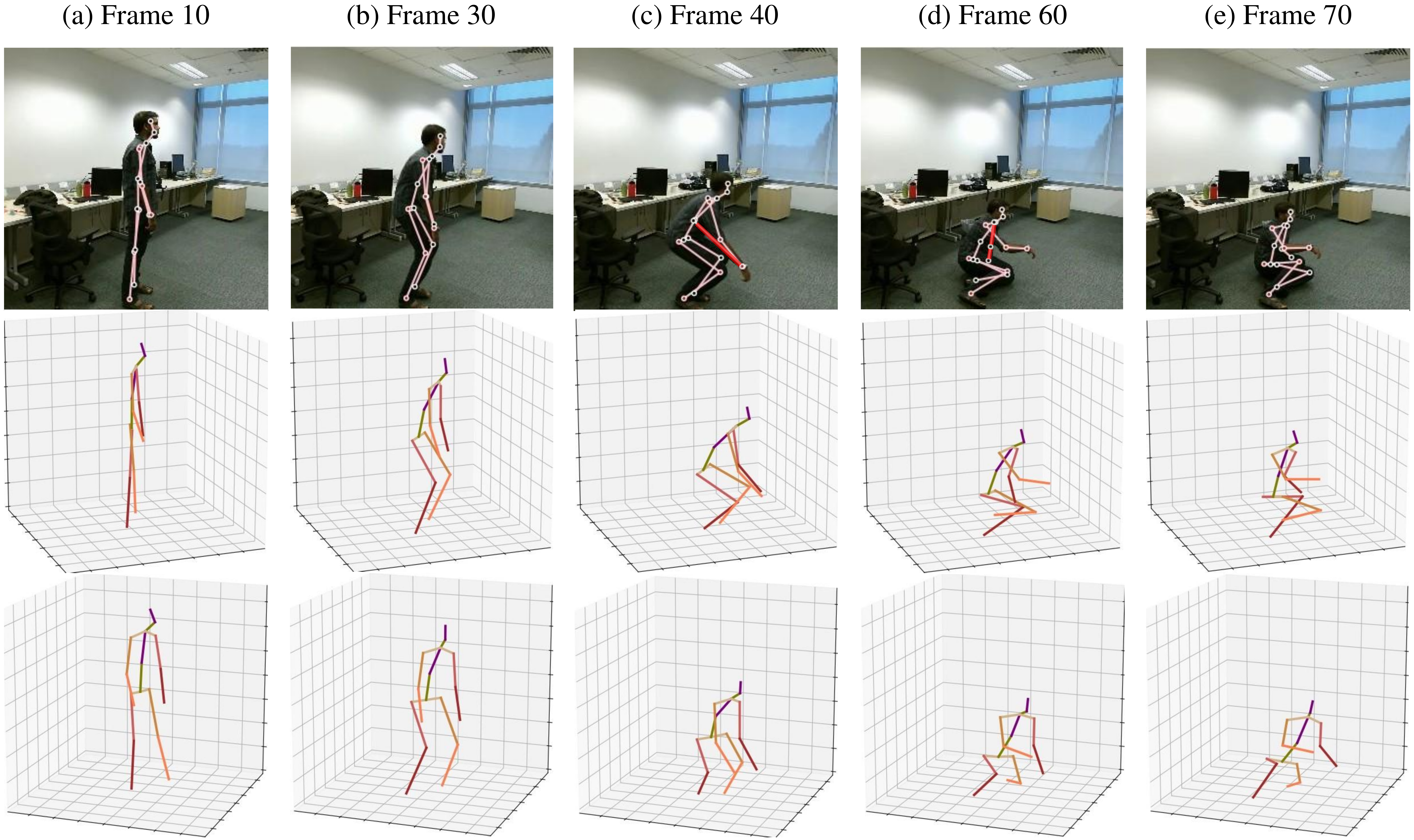}
	\caption{Pose estimation reconstruction under depth ambiguities, self-occlusion, and biased 2D poses. Row 1: 2D pose estimation, where red is the prediction with errors. Row 2, 3: our reconstructions from two different perspectives. 
	} \label{fig:occlusion}
\end{figure}

Estimating 3D poses from 2D keypoints remains an ill-posed problem due to: (i) depth ambiguity: caused by a many-to-one 3D-to-2D pose mapping as shown in Fig. \ref{fig:occlusion} (a); (ii) self-occlusions: caused under certain human poses as shown in Fig. \ref{fig:occlusion} (a-e)); and (iii) prediction errors: caused by inaccurate human pose models as shown in Fig. \ref{fig:occlusion} (c,d).

To mitigate depth ambiguity, some works \cite{drover2018can, wandt2019repnet} introduce weakly supervised methods to regular the rationality of the generated 3D structure. For example, Drover \et \cite{drover2018can} utilizes an adversarial framework to impose a prior on the 3D structure via random 2D-to-3D projections. Even so, self-occlusion still remains difficult to solve as well as jittery motions in estimated video.
To address these problems, temporal modeling has used joint-coordinated vectors in sequence-to-sequence models to generate smoother motion \cite{rayat2018exploiting,pavllo20193d}. However, the vector representation of joint sequences lacks expressivity for spatial relations, which is critical to mitigate depth ambiguities and self-occlusions.
To make full use of  spatio-temporal information, Cai \et \cite{cai2019exploiting} designs a local-to-global networks to construct spatial configurations and temporal consistencies, which takes 2D keypoints sequence as a spatio-temporal graph. But, The use of graph convolutional networks to encode temporal relationships cannot effectively model long-term dependencies. This network architecture is also limit to input fixed length.

Despite considerable progress in 2D-to-3D video-based methods, none of them integrate the following significant characteristics:

(i) Extract more informative contextual spatio-temporal information from the hierarchical structure of 2D keypoint sequences and posture semantics to resolve depth ambiguities, mitigate self-occlusions, and smoothen motion.

(ii) Flexible input frame length handling, noting that videos effectively exhibit varying length sequences.

(iii) Real-time estimation of 3D poses without redundant intermediate frame calculations. Real-time estimation facilitates the downstream combination of high-semantic visual tasks; such as combining skeleton-based action recognition with human-robot interaction in real-time.

\begin{figure*}[thbp]
	\centering
	\includegraphics[width=1.0\textwidth]{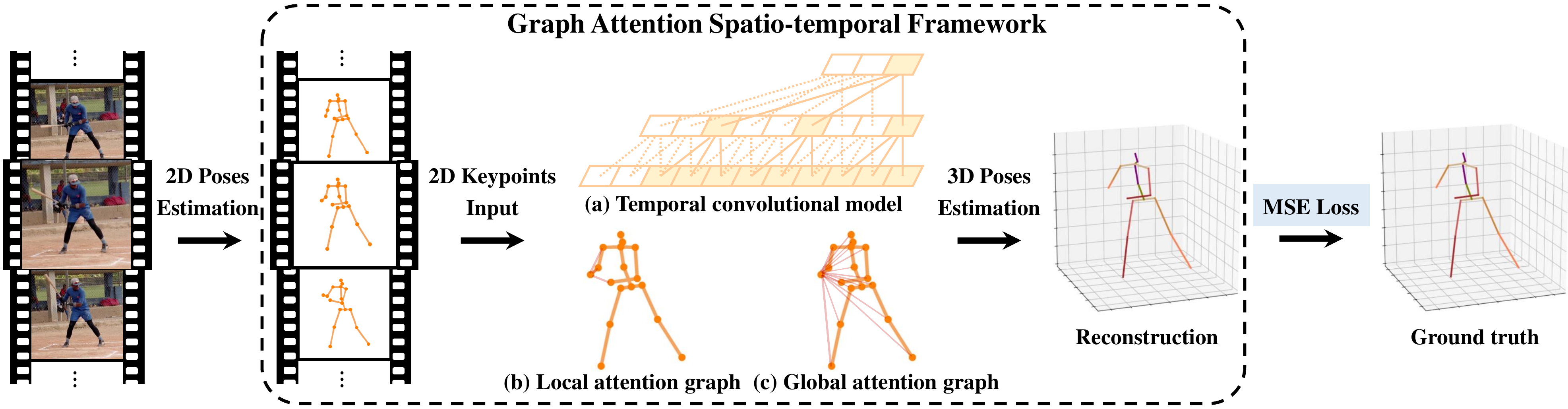}
	\caption{Schematic overview of the GAST-Net framework. The input consists of consecutive 2D pose estimates from RGB images. The output is a sequence of reconstructed 3D poses from the corresponding 2D keypoints. GAST-Net synergistically interleaves 3 components: (a) a dilated temporal convolutional model (with 2D keypoint sequences as input (bottom) and 3D pose estimates as output (top)) with (b) a set of local attention mechanisms for visualized joints (\emph{i.e.} the right-wrist) including local kinematic dependencies and symmetric relations, and (c) a global attention mechanism that informs about posture semantics.}
	\label{fig:framework}
\end{figure*}
These challenges inspire us to study richer spatio-temporal representations and flexibly interleave spatial and temporal information. To this end, we contribute an interleaved graph-attention spatio-temporal network that better learns three aspects of human kinematic constraints via graph attention blocks and leverages dilated convolutions to model long-term temporal contexts. The graph attention block learns skeletal joint symmetries, local kinematic relations in distal joints, and global postural joint semantics. The dilated temporal convolutional networks (TCNs) can flexibly capture varying sequences and work with causal convolutions to achieve real-time pose estimation\cite{pavllo20193d,bai2018an} (see Fig. \ref{fig:framework} (a)). For single-frame scenarios, dilated convolutions can be replaced with strided convolutions to quickly inference without need to retrain a new model \cite{pavllo20193d}. In our works, we understand spatial and temporal data to be heterogeneous. As such we treat them independently but interleave them in a synergistic manner allowing us to leverage the benefits of TCNs.

\begin{figure*}[ht]
	\centering
	\includegraphics[width=0.98\textwidth]{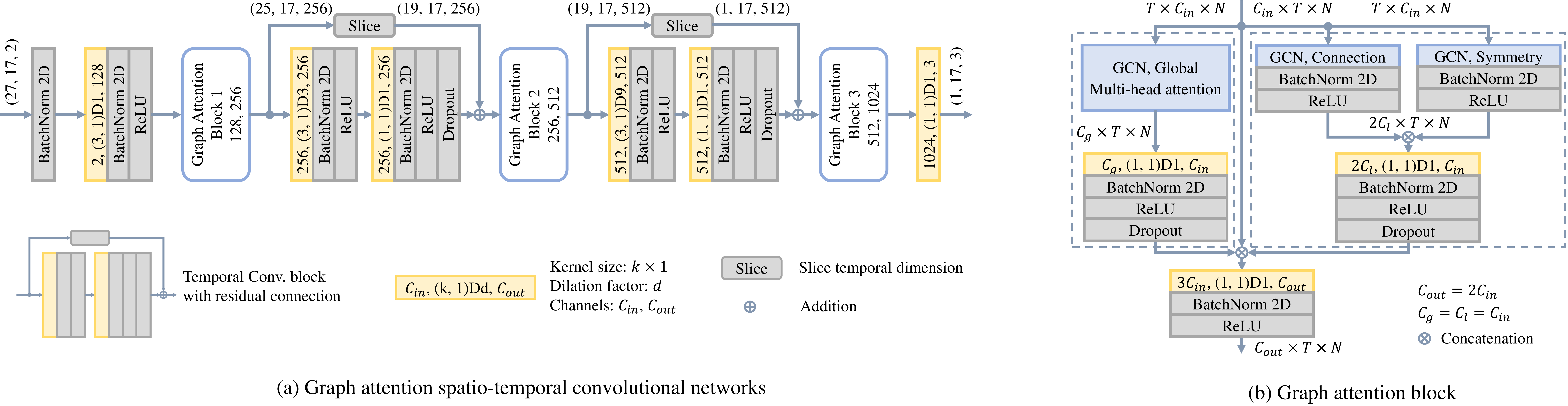}
	\caption{(a) An instantiation of GAST-Net for 3D pose estimation. The GAST-Net consists of 2 Temporal Convolution Blocks and 3 Graph Attention Blocks. Given a 2D pose sequence, the output is a sample 1-frame prediction. Dimensions are enclosed in parenthesis: \emph{e.g.} (27, 17, 2) denotes a receptive field of 27 frames, 17 joints, and 2 channels. (b) The graph attention block architecture. The left dotted box indicates the local graph attention layer. The right dotted box indicates the global graph attention layer. The layer outputs is concatenated followed by a 2D convolution layer before outputting the spatio-temporal features.}
	\label{fig:gast_net}
\end{figure*}

With regards to temporal modeling, we base our design on the dilated temporal convolutions of \cite{pavllo20193d}, but extend it to tackle three-dimensional spatio-temporal sequences.
With regards to spatial modeling, local spatial features from local connections and symmetries are modeled via graph convolution networks (GCNs) and referred to as ``Local Attention Graph's'' in our system (see Fig. \ref{fig:framework} (b)). 
For global spatial features, we draw inspiration from \cite{shi2019two} and leverage graph attention networks \cite{hamilton2017inductive} to express posture semantics with data-driven learning. These blocks are referred to as ``Global Attention Graph's'' and depicted in Fig. \ref{fig:framework} (c). 
The graph attention block effectively express the hierarchical symmetrical structure of the human body and adaptively extracts global semantic information over time. Particularly, local- and global-spatial blocks are interleaved with temporal blocks to effectively extract and fuse spatio-temporal features of 2D keypoint sequences (see Fig. \ref{fig:gast_net}).

\section{Related Works}
\subsection{2D-to-3D Pose Estimation}
Since Martinez \emph{et al.} \cite{martinez2017simple} proposed a simple and effective linear layer to lift 2D joint locations to 3D positions, many works \cite{dabral2018learning,wandt2019repnet,wang2019generalizing} have sought to generate accurate 3D pose estimation from underlying 2D keypoints. Wandt \emph{et al.} \cite{wandt2019repnet}, proposed a semi-supervised approach to solve overfitting by projecting a generated 3D pose back to the 2D image and comparing it with the ground truth. Wang \emph{et al.} \cite{wang2019generalizing}, designed a novel stereo network with a geometric search scheme to generate a high quality 3D pose in the wild without the need of indoor 3D input. Even so, generating accurate 3D poses from a single image is an ill-posed problem. Recent works have exploited  temporal information to obtain more robust and smooth 3D poses \cite{dabral2018learning,rayat2018exploiting,pavllo20193d,cai2019exploiting,lee2018propagating}. For instance, Hossain \et \cite{rayat2018exploiting} proposed a 2-layered normalized LSTM network with residual connections that first encode 2D poses into a fixed feature vector and then decode it to a 3D pose. But, encoding the two-dimensional poses into a one-dimensional vector ignores the expression of the spatial configuration of 2D poses. Other recent works incorporate spatial configuration constraints and temporal information to estimate 3D poses \cite{cai2019exploiting, wang2020motion}. Wang \cite{wang2020motion} \et designed a U-shaped graph convolutional network to aggregate long-range information through temporal pooling operations. Cai \cite{cai2019exploiting} \et exploited graph pooling and graph upsampling to process and consolidate features across scales. However, their local-to-global network architectures are limit to embed fixed-length spatio-temporal sequences.

\subsection{Spatio-Temporal Graph}
GCNs generalize convolutions to graph-structured data and are roughly classified into  spectral-based and spatial-based categories \cite{kipf2016semi,zhao2019semantic,velivckovic2017graph}. Spatial-based GCNs are more relevant to our work. Our spatial network uses both GCNs proposed by \cite{zhao2019semantic} and \cite{velivckovic2017graph} to obtain local and global features of each joint. Due to the outstanding performance of GCNs in non-European data, there are many recent works also modeling skeleton sequence as spatio-temporal graphs to understand human tasks including skeleton-based action recognition \cite{shi2019two,yan2018spatial} and motion prediction \cite{li2020dynamic}. Our approach bears some similarity with adaptive graph convolutional blocks \cite{shi2019two} that extends the topology of the graph and combines common convolutional networks to integrate the spatio-temporal information. But our works has four distinct features: (i) instead of setting the local spatial configuration to three subsets based on gravity, our approach aims to model the symmetrical hierarchy of the human body as well as the kinematic joint constraints; (ii) our local and global adjacency matrices are applied to different graph convolutions to explicitly extract diverse spatial semantics; (iii) the dilated convolution is used to effectively model long-term temporal information; and (iv) as with the inception module, we exploit concatenation to better integrate the three-steam spatio-temporal features.

\section{Approach}\label{sec:GAST}
Given a sequence of 2D pose predictions from videos, our goal is to output a sequence of 3D coordinates based on a root joint---the pelvis (See Fig. \ref{fig:ske_local_mat} (a)). In this section, we introduce our interleaved graph attention spatio-temporal network. The temporal component is designed from dilated TCNs to tackle long-term patterns (Sec. \ref{subsec:TempConvs}). As for the spatial components, we have a local spatial attention network (Sec. \ref{subsec:localGCNs}) to model the hierarchical and symmetrical structure of the human skeleton and a global spatial attention network (Sec. \ref{subsec:globalGCNs}) to adaptively extract global semantic information to better encode the human body's spatial characteristic. 
Fig. \ref{fig:gast_net} (a) depicts an instantiation of the proposed framework with a receptive field size of 27 frames, whilst Fig. \ref{fig:gast_net} (b) depicts the graph attention block which is composed of local and global spatial blocks. 
\subsection{Temporal Convolutional Network}\label{subsec:TempConvs}
The original temporal dilated convolutional model  \cite{pavllo20193d} consists of an input layer, an output layer and $B$ temporal convolutional blocks that flexibly control the receptive field by setting the kernel size and the dilation factor of the convolution. Each block first performs a 1D convolution with kernel size $k$ and dilation factor $d = k^B$, followed by a convolution with kernel size 1. The main difference compared to \cite{pavllo20193d} is that we represent the input 2D pose sequence as a three-dimensional vector $(T, N, C)$, where $T$ is the number of receptive fields, $N$ is the number of joints in each frame, and $C$ is the number of coordinate dimensions $(x,y)$. To save the spatial information across time steps, we replace the original 1D convolution with 2D convolutions for a kernel size of $ k \times 1 $. Simultaneously, each batch normalization is changed to 2D, and it is added at the beginning to normalize the input data. The dropout is only employed at the second convolution layer of blocks to improve generalization. Fig. \ref{fig:gast_net} shows an instantiation of GAST-Net for a receptive field size of 27 frames with $B=2$ blocks. Note that according to the network characteristics of TCNs, our proposed model can train under varying numbers of long-sequence receptive fields as needed.
\subsection{Local Attention Graph}\label{subsec:localGCNs}
At any given frame in time, 2D keypoints represent the joints of the human skeleton. The skeleton is naturally represented by an undirected graph with joints as nodes and human-links as edges. We construct a skeleton graph of 2D keypoints for a given frame based on the SemGCN proposed by Zhao \cite{zhao2019semantic}. We define the skeletal 2D poses as a graph $\mathcal{G} = (\mathcal{V}, \mathcal{E})$, where $\mathcal{V}$ is the set of $N$ nodes and $\mathcal{E}$ edges. $X = \{x_{1}, x_{2}, \dots, x_{N} \mid x_{i} \in \mathbb{R}^{1 \times C} \}$ is the set of node features with $C$ channels. The structure of the graph can be initialized by a first-order adjacency matrix $A \in \mathbb{R}^{N \times N}$ that indicates the existing connections between joints and an identity matrix $I$ indicating self-connections. $\widetilde{A} = (A + I)$ expresses the convolutional kernel in GCNs. According to the definition of SemGCN, given the node features of the $l$-th layer, the output features of the subsequent layer are obtained through the following convolution:
\begin{equation}
X^{(l+1)} = \rho ( M \odot \widetilde{A} ) X^{\left( l \right)} W, 
\label{eqtn:convKernel}
\end{equation}
where $W \in \mathbb{R}^{C_{l} \times C_{l+1}}$ is a learnable matrix used to transform output channels, $M \in \mathbb{R}^{N \times N}$ is a learnable mask matrix, $\odot$ is an element-wise multiplication operation, and $\rho$ is a Softmax nonlinearity that normalizes the contribution of the features of a node to a corresponding neighboring node in a graph. 

By introducing a set of mask matrices $M_{c} \in \mathbb{R}^{N \times N}$ for the channels of the output node features, Eqtn. \ref{eqtn:convKernel} can be extended to:
\begin{equation}
X^{(l+1)} = \mathop{\left| \right|} \limits_{c=1}^{C_{l+1}} \rho ( M_{c} \odot \widetilde{A} ) X^{\left( l \right)} w_{c}, 
\label{eqtn:maskMat}
\end{equation}
where $\parallel$ denotes a channel-wise concatenation and $w_{c}$ is the $c$-th row of matrix $W$.

Eqtn. \ref{eqtn:maskMat} jointly learns the unique semantics across neighboring nodes. However, and very notably, this first-order neighbor representation poorly models (i) the symmetrical structure of a torso-centered human body and (ii) kinematic constraints in the human body. 
Thus, we propose that structural knowledge pertaining to symmetrical functions in the human body is explicitly considered. 
Furthermore, another reason why first-order neighbor representations struggle to model human spatial relations is that joint constraints are confined to first-order neighboring joints. More precisely, distal joints like the wrist, ankle, and head, located at the end of the kinematic chain, only have one first-order neighboring joint. As such, their position in space are not effectively located due to the first-order neighborhood. Such joints are the largest single source of modeling errors \cite{rayat2018exploiting}. Nonetheless, we exploit the relationship across sub-segments of the kinematic chain; that is, the lower limbs (ankle-knee-hip), upper limbs (wrist-elbow-shoulder), and the axial-body (head-neck-thorax) to mitigate  location ambiguity.

\begin{figure}[ht]
	\centering
	\includegraphics[width=0.46\textwidth]{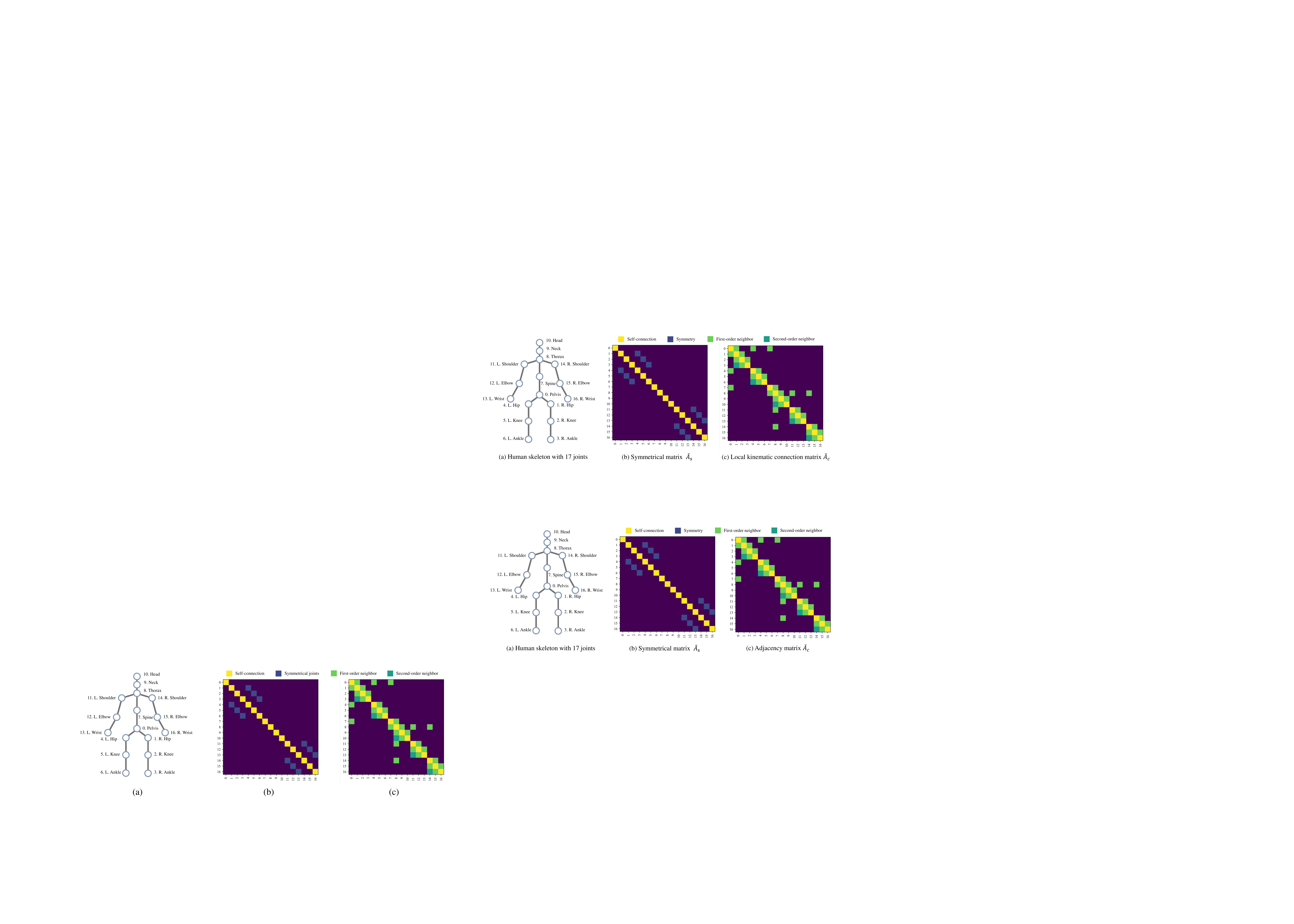}
	\caption{(a) A human skeleton graph with 17 joints; (b) A symmetrical matrix $\widetilde{A}_{s}$; (c) An adjacency matrix $\widetilde{A}_{c}$.}
	\label{fig:ske_local_mat}
\end{figure}

Based on the aforementioned limitations, we design two novel convolution kernels: 
(i) a symmetric matrix $\widetilde{A}_{s}$ (See Fig. \ref{fig:ske_local_mat} (b)) that encodes the human skeleton symmetrical structure for joints that have a symmetrical counterpart (\emph{i.e.} limb joints).
(ii) an adjacency matrix $\widetilde{A}_{c}$ (See Fig. \ref{fig:ske_local_mat} (c)) that explicitly encodes first- and second-order (kinematic) connections for distal joints (\emph{i.e.} ankle-knee, ankle-hip). The rest of the nodes are only modeled through first-order connections. 

Note that each of these two convolution kernels are applied to two distinct GCNs; where each GCNs is followed by batch normalization and rectified linear units as shown in the right dotted box of Fig. \ref{fig:gast_net} (b).

\subsection{Global Attention Graph}\label{subsec:globalGCNs}
The relationship across disconnected joints, those that exist across sub-segments of the skeleton (\eg wrist-ankle), play a key role in encoding global postural and constraint information (think running). As such, disconnected joint representation aid in addressing depth ambiguities and occlusions. To adaptively and effectively encode non-local relationships, we propose a global end-to-end GCN with a multi-head attention mechanism that extends the mechanism introduced in Eqtn. \ref{eqtn:maskMat} from first-order relationships to global relationships. The global attention mechanism is first presented in Eqtn. \ref{eqtn:globalGraph} and then explained in further detail.
\begin{equation}
X^{(l+1)} = \mathop{\left| \right|} \limits_{k=1}^{K}  \left( B_{k} + C_{k} \right) X^{(l)} W_{k} 
\label{eqtn:globalGraph}
\end{equation}
\noindent where, $K$ is the number of attention heads, $B_{k} \in \mathbb{R}^{N \times N}$ is an adaptive global adjacency matrix, $C_{k} \in \mathbb{R}^{N \times N}$ is a learnable global adjacency matrix, and $W_{k} \in \mathbb{R}^{C_{l} \times (C_{l}/K)}$ is a transformed matrix. In this work we set $K = 4$ parallel attention heads. Next, we discuss the redefined adjacency matrix $B_{k}$ and the global adjacency matrix $C_{k}$ in detail.

\textbf{$B_{k}$} expresses a data-dependent matrix which learns a unique graph for each node. We adopt the attention coefficient function proposed by \cite{velivckovic2017graph} to determine whether a connection exists between nodes and how strong the connection is. That is, given two node features $x_{i}$ and $x_{j}$, we first apply two embedding functions $\theta$ and $\phi$ to downsample the features of each node from $C_{l}$ to $C_{l}/K$ channels. Since the number of channels for each node is reduced, the total computational cost for multi-attention is similar to that of single-headed attention with full channels. Then we concatenate the two embedded features, and compute their dot product with a weight vector $w_{f}$ to produce a scalar output. To facilitate coefficient comparisons across nodes, the scalar outputs are normalized by the softmax function. The operation is presented in Eqtn. \ref{eqtn:attn_coef}:
\begin{equation}
\alpha_{ij}  = 
\frac{
	e^{\sigma \left( w_{f} \cdot \lbrack \theta(x_{i}) \Arrowvert \phi(x_{j}) \rbrack \right) } 
}
{
	\sum_{k=1}^{N} e^{\sigma \left( w_{f} \cdot \lbrack \theta(x_{i}) \Arrowvert \phi(x_{k}) \rbrack \right) }
},
\label{eqtn:attn_coef}
\end{equation}
where $\theta$ and $\phi$ are convolutions with the kernel size 1; $[ \cdot \Arrowvert \cdot]$ denotes concatenation, and $\sigma$ denotes a LeakyReLU nonlinearity with negative input slope $\alpha=0.2$.

\textbf{$C_{k}$} is a learnable adjacency matrix, inspired by \cite{shi2019two}, with an initialization value of zero. The value of $C_{k}$ is not limited to special node features like $B_{k}$, which is updated during the training process. The elements of \textbf{$C_{k}$} are arbitrary. They indicate the existence and strength of connections between two joints. It plays a similar role to the attention mechanism performed by $M_c$ in Eqtn. \ref{eqtn:maskMat}. However, note how in Eqtn. \ref{eqtn:maskMat},  $M_c$ is dot multiplied with $\widetilde{A}$, and if any of the elements in $\widetilde{A}$ is 0, the product will always be 0 irrespective the value of $M_c$. Thus, no new connections can be created in the original physical graph. From this perspective, \textbf{$C_{k}$} is more flexible than $M_c$.
\section{Datasets}
We evaluate our method on two publicly available datasets: Human3.6M \cite{ionescu2013human3} and HumanEva-I \cite{sigal2010humaneva}. Human3.6M captures data through four synchronized cameras at 50 Hz and contains 3.6 million video frames with 11 professional subjects performing 15 daily activities (\emph{i.e.} walking and sitting). Following previous methods \cite{martinez2017simple,rayat2018exploiting,pavllo20193d}, we employ subjects 1, 5, 6, 7, 8 for training and 9, 11 for testing. HumanEva-I, is a much smaller dataset and captures data through three camera views at 60 Hz. Following \cite{rayat2018exploiting,pavllo20193d}, the time-series data from three actions (walk, jog, box) is split between training and testing. 

We use two common evaluation protocols in our experiments. Protocol \#1 calculates the mean per joint positioning error (MPJPE) between the ground truth and the predicted 3D coordinates across all cameras and joints. Protocol \#2 employs a rigid alignment (Procrustes analysis) with the ground truth before calculating the mean per joint positioning error (P-MPJPE).

\section{Experiments}
\subsection{Training \& Inference}

GAST-Net was trained with receptive fields of sizes 9, 27, 81 and 243 to verify the effectiveness of our model architecture. To make the model lightweight, for networks with receptive fields of 9 and 27, we increase the number of output channels of the first dilated convolutional layer to 128, while the network with 81 and 243 receptive fields is set to 64 and 32 channels respectively. Note that our loss function only computes the MPJPE between the predicted 3D location and ground truth without using any tricks (\emph{i.e.} motion constraint and pose regulation).

\subsubsection{Training \& Inference Strategy}
To train the proposed model, we use Pavllo's optimized training strategy for single-frame predictions instead of the layer-by-layer implementation \cite{pavllo20193d}.
For single-frame scenarios, dilated convolutions are known to waste a large number of computations. To reduce the inefficiency, dilated convolutions are replaced with strided convolutions. At inference, we switch and consider the entire video sequence. We change from the optimized training strategy to the layer-by-layer implementation to make faster predictions. 

\subsubsection{Implementation Details}
Note that the different datasets have different joint setups. In Human 3.6M we predict poses using a 17-joint skeleton and in HumanEva-I, we use 15-joints. Also, we noted that high frame rates lead to information redundancy that negatively affects the encoding of global semantics over time. For this reason, we decided to downsample the Human3.6M dataset from 50 FPS to 10 FPS. On the other hand, as the duration of videos in the HumanEva-I dataset is short, no downsampling is performed here. With regards to real-time estimation, long video durations are not suitable for fast estimation; as such we do not perform downsampling for the 243 receptive field model. Finally, we adopt horizontal flip augmentation at both training and testing time.

We implement our method with the PyTorch framework and train end-to-end. For Human3.6M, we optimize with Amsgrad with a mini-batch size of $b=128$, and train for 80 epochs. The learning rate starts at 0.001 and then applies a learning shrink factor $\alpha = 0.95$ in each epoch. The dropout rate $p$ in each dropout layer is set to 0.05. For HumanEva-I, we use $b=32$, $\alpha=0.98$, $p=0.5$, and train for 200 epochs.
\subsection{Ablation Studies}
In our ablation studies, as with those in \cite{cai2019exploiting, pavllo20193d}, we will use 2D poses detected from Cascaded Pyramidal Networks (CPNs) \cite{chen2018cascaded} in our 27 receptive fields model.
\subsubsection{Effects of Spatial Semantics}
\begin{table}[tbp]
	\caption{Ablation study on different spatial semantics in our network architecture on Human3.6M under both protocols.}
	\label{table:1}
	\begin{center}
		\begin{tabular}{l|cc}
			\toprule[1pt]
			Method (T=27, CPN)                       & MPJPE (mm) & P-MPJPE (mm)   
			\\ \midrule[0.5pt]
			Baseline                                 & 60.9      & 50.5          \\
			+ Local $GCN_{s}$ with $\widetilde{A}_{c}$ & 55.4      & 44.0          \\
			+ Local $GCN_{s}$ with $\widetilde{A}_{s}$ & 51.9      & 40.9          \\
			+ Global $GCN_{s}$  with $B_{k}$         & 47.3      & 36.2          \\
			+ Global $GCN_{s}$  with $C_{k}$         & 46.2      & 36.0         \\ \bottomrule[1pt]
		\end{tabular}
	\end{center}
\end{table}
\begin{table}[tbp]
	\caption{Ablation study on the sensitivity of spatial semantics in our model on Human3.6M under Protocol \# 1.}
	\label{table:2}
	\begin{center}
		\begin{tabular}{l|c|c}
			\toprule[1pt]
			Method (T=27, CPN)                                & MPJPE(mm) & $\Delta$   
			\\ \midrule[0.5pt]
			Ours w/o Local $GCN_{s}$ with $\widetilde{A}_{c}$ & 47.6      & 1.4        \\
			Ours w/o Local $GCN_{s}$ with $\widetilde{A}_{s}$ & 47.4      & 1.2        \\ 
			\midrule[0.5pt]
			Ours (GAST-Net)                                   & 46.2      & -          
			\\ 
			\bottomrule[1pt]
		\end{tabular}
	\end{center}
\end{table}
We perform ablation studies to analyze the effect of spatial semantics in our networks architecture as shown in Table \ref{table:1}. As the baseline, we build a plain GAST-Net comprised of TCNs and first-order SemGCNs to regress 2D keypoints to 3D poses. We then add different semantic GCNs one-by-one to conduct ablation studies. The semantics consist of: 
(i) local kinematic relations $\widetilde{A}_{c}$, 
(ii) symmetric relations $\widetilde{A}_{s}$, 
(ii) global adaptive matrix $B_{k}$,  and 
(iv) global learnable matrix $C_{k}$. 
We see that as we consider additional local-to-global postural constraints, the performance improves steadily. The largest improvements come from local kinematic connections, symmetry, and the global adaptive matrix. These spatial constraints exactly express a hierarchical and symmetrical human structure and conveys global posture semantics, which better reconstructs valid 3D poses. These results support the significance of rich spatial semantics in 3D pose estimation.
\subsubsection{Sensitivity Analysis of Spatial Semantics}
Global graph matrices ($B_{k}$ and $C_{k}$) encompass both local and symmetric joint relations. We wish to explore the contributions of each of these configurations under the global analysis. To this end, we study the effect of removing local kinematic connections $\widetilde{A}_{c}$ and symmetry $\widetilde{A}_{s}$ separately on the Human3.6M as shown in Table \ref{table:2}. 
The study reveals that removing local connections and symmetry increase errors by 1.4mm and 1.2mm respectively. From this we conclude that explicitly embedding local connections and symmetrical prior knowledge is indispensable and complementary to global semantics in generating more accurate 3D poses.
\subsection{Quantitative Results}
\begin{table*}[t]
	\centering
	\caption{Quantitative comparisons of MPJPE in millimeters between the estimated pose and the ground-truth (GT) on the Human3.6M under Protocol \#1 and Protocol \#2. $T$ denotes the number of receptive fields, $(\dag)$ indicates the use of pose refinement and spatio-temporal information. Best in bold, second best underlined.}
	\label{table:3}
	\resizebox{0.97\textwidth}{!}{
		\begin{tabular}{l|ccccccccccccccc|c}
			\toprule[1pt]
			\multicolumn{1}{l|}{Protocol \#1} & Dir.          & Disc.         & Eat           & Greet         & Phone         & Photo         & Pose          & Purch.        & Sit           & SitD.         & Smoke         & Wait          & WalkD.        & Walk          & WalkT.        & Avg           \\ \midrule[0.5pt]
			Hossain \cite{rayat2018exploiting}            & 48.4          & 50.7          & 57.2          & 55.2          & 63.1          & 72.6          & 53.0          & 51.7          & 66.1          & 80.9          & 59.0          & 57.3          & 62.4          & 46.6          & 49.6          & 58.3          \\
			Cai \cite{cai2019exploiting} ($\dag$)         & 44.6          & 47.4          & 45.6          & 48.8          & 50.8          & 59.0          & 47.2          & 43.9          & 57.9          & 61.9          & 49.7          & 46.6          & 51.3          & 37.1          & 39.4          & 48.8          \\
			Pavllo \cite{pavllo20193d}                    & 45.2          & 46.7          & 43.3          & 45.6          & 48.1          & 55.1          & 44.6          & 44.3          & 57.3          & 65.8          & 47.1          & 44.0          & 49.0          & 32.8          & 33.9          & 46.8          \\
			Lin \cite{lin2019trajectory}                  & 42.5          & {\ul 44.8}    & 42.6          & {\ul 44.2}    & 48.5          & 57.1          & {\ul 42.6}    & \textbf{41.4} & 56.5          & 64.5          & 47.4          & {\ul 43.0}    & 48.1          & 33.0          & 35.1          & 46.6          \\
			Liu \cite{liu2020attention}                   & {\ul 41.8}    & {\ul 44.8}    & {\ul 41.1}    & 44.9          & 47.4          & {\ul 54.1}    & 43.4          & {\ul 42.2}    & {\ul 56.2}    & 63.6          & \textbf{45.3} & 43.5          & \textbf{45.3} & {\ul 31.3}    & 32.2          & 45.1          \\
			Wang \cite{wang2020motion} ($\dag$)           & \textbf{40.2} & \textbf{42.5} & 42.6          & \textbf{41.1} & {\ul 46.7}    & 56.7          & \textbf{41.4} & 42.3          & {\ul 56.2}    & {\ul 60.4}    & 46.3          & \textbf{42.2} & {\ul 46.2}    & 31.7          & {\ul 31.0}    & \textbf{44.5} \\ \midrule[0.5pt]
			Ours  (T=243 CPN causal)          & 45.5          & 48.4          & 43.9          & 48.3          & 49.3          & 57.6          & 45.0          & 45.8          & 57.3          & 61.4          & 49.3          & 45.3          & 49.6          & 33.7          & 33.4          & 47.7          \\
			Ours (T=243 CPN)                  & 43.3          & 46.1          & \textbf{40.9} & 44.6          & \textbf{46.6} & \textbf{54.0} & 44.1          & 42.9          & \textbf{55.3} & \textbf{57.9} & {\ul 45.8}    & 43.4          & 47.3          & \textbf{30.4} & \textbf{30.3} & {\ul 44.9}    \\ \midrule[0.5pt]
			Ours (T=243 GT)               & 30.5          & 33.1          & 27.6          & 31.0          & 31.8          & 37.0          & 33.2          & 30.0          & 35.7          & 37.7          & 31.4          & 29.8          & 31.7          & 24.0          & 25.7          & 31.4          \\ \bottomrule[1pt]
			
			\specialrule{0em}{1pt}{1pt}
			Protocol \#2            & Dir.          & Disc.         & Eat           & Greet         & Phone         & Photo         & Pose          & Purch.        & Sit           & SitD.         & Smoke         & Wait          & WalkD.        & Walk          & WalkT.        & Avg           \\ \midrule[0.5pt]
			Hossain \cite{rayat2018exploiting}            & 35.7          & 39.3          & 44.0          & 43.0          & 47.2          & 54.0          & 38.3          & 37.5          & 51.6          & 61.3          & 46.5          & 41.4          & 47.3          & 34.2          & 39.4          & 44.1          \\
			Cai \cite{cai2019exploiting} ($\dag$)         & 35.7          & 37.8          & 36.9          & 40.7          & 39.6          & 45.2          & 37.4          & 34.5          & 46.9          & 50.1          & 40.5          & 36.1          & 41.0          & 29.6          & 33.2          & 39.0          \\
			Pavllo \cite{pavllo20193d}    & 34.1          & 36.1          & 34.4          & 37.2          & 36.4          & 42.2          & 34.4          & 33.6          & 45.0          & 52.5          & 37.4          & 33.8          & 37.8          & 25.6          & 27.3          & 36.5          \\
			Lin \cite{lin2019trajectory}                  & 32.5          & 35.3          & 34.3          & 36.2          & 37.8          & 43.0          & {\ul 33.0}    & {\ul 32.2}    & 45.7          & 51.8          & 38.4          & 32.8          & 37.5          & 25.8          & 28.9          & 36.8          \\
			Liu \cite{liu2020attention}                   & {\ul 32.3}    & {\ul 35.2}    & \textbf{33.3} & {\ul 35.8}    & {\ul 35.9}    & \textbf{41.5} & 33.2          & 32.7          & 44.6          & 50.9          & 37.0          & {\ul 32.4}    & 37.0          & 25.2          & 27.2          & 35.6          \\
			Wang \cite{wang2020motion} ($\dag$)           & \textbf{31.8} & \textbf{34.3} & 35.4          & \textbf{33.5} & \textbf{35.4} & {\ul 41.7}    & \textbf{31.1} & \textbf{31.6} & {\ul 44.4}    & 49.0          & \textbf{36.4} & \textbf{32.2} & \textbf{35.0} & {\ul 24.9}    & \textbf{23.0} & \textbf{34.5} \\ \midrule[0.5pt]
			Ours (T=243 CPN causal) & 34.9          & 37.5          & 34.9          & 38.3          & 37.4          & 44.0          & 34.4          & 34.6          & 45.1          & {\ul 48.0}    & 49.3          & 34.8          & 37.7          & 26.2          & 27.1          & 36.9          \\
			Ours (T=243 CPN)        & 32.7          & 36.2          & {\ul 33.4}    & 36.5          & 36.0          & \textbf{41.5} & 33.6          & 33.1          & \textbf{44.1} & \textbf{46.8} & {\ul 36.7}    & 33.1          & {\ul 35.8}    & \textbf{24.2} & {\ul 24.8}    & {\ul 35.2}    \\ \midrule[0.5pt]  
			Ours (T=243 GT)               & 23.4          & 26.8          & 21.6          & 25.0          & 24.5          & 28.9          & 25.6          & 23.4          & 29.0          & 30.5          & 26.1          & 23.3          & 25.8          & 20.3          & 21.7          & 25.1          \\ \bottomrule[1pt]
		\end{tabular}
	}
\end{table*}
\begin{table}[t]
	\centering
	\caption{Comparison on HumanEva-I under protocol \#2. Best in bold, second best underlined. Note that the high error on S3's ``Walk'' is due to corrupted mocap data.}
	\label{table:4}
	\resizebox{0.49\textwidth}{!}{
		\begin{tabular}{l|ccc|ccc|ccc}
			\toprule[1pt]
			\multicolumn{1}{l|}{\multirow{2}{*}{Protocol \#2}} & \multicolumn{3}{c|}{Walk}                    & \multicolumn{3}{c|}{Jog}                       & \multicolumn{3}{c}{Box}                      \\ \midrule[0pt] \cline{2-10} \midrule[0pt]
			\multicolumn{1}{c|}{}                        & S1            & S2           & S3            & S1            & S2            & S3            & S1            & S2            & S3            \\ \midrule[0.5pt]
			Martinez \cite{martinez2017simple}                           & 19.7          & 17.4         & 46.8          & 26.9          & 18.2          & 18.6          & -             & -             & -             \\
			Pavlakos \cite{pavlakos2018ordinal}                          & 18.8          & 12.7         & \textbf{29.2} & 23.5          & 15.4          & 14.5          & -             & -             & -             \\
			Lee \cite{lee2018propagating}                                & 18.6          & 19.9         & {\ul 30.5}    & 25.7          & 16.8          & 17.7          & 42.8          & 48.1          & 53.4          \\
			Pavllo \cite{pavllo20193d}                                   & {\ul 13.9}    & {\ul 10.2}   & 46.6          & {\ul 20.9}    & {\ul 13.1}    & {\ul 13.8}    & {\ul 23.8}    & {\ul 33.7}    & {\ul 32.0}    \\ \midrule[0.5pt]
			Ours (T=27 CPN)                                  & \textbf{13.7} & \textbf{9.2} & 46.2          & \textbf{20.1} & \textbf{12.5} & \textbf{12.7} & \textbf{21.8} & \textbf{27.8} & \textbf{27.0} \\ \bottomrule[1pt]
		\end{tabular}
	}
\end{table}
\subsubsection{Comparison with State-of-the-Art}
With respect to the Human3.6M dataset, we train using CPNs \cite{chen2018cascaded} to detect 2D poses for fair comparison as this is the most commonly used detector in the compared works. Table \ref{table:3} shows the performance of our 243 receptive field model compared to state-of-the-art (SOTA) results. Our method achieves competitive performance on human3.6M under both protocols. Note that Wang \emph{et al.} \cite{wang2020motion} not only exploited spatial and temporal information but also adopted pose refinement and motion loss to regular reconstructed 3D poses. In our work, we only model spatio-temporal information through a simple network and use a common MPJPE loss without using any bells and whistles. In addition, we also report the results when using ground truth 2D poses, which yield approximately 13.5mm improvements in MPJPE.

With respect to HumanEva-I dataset, which is comprised of videos with much smaller durations compared with those in Human3.6M, we choose to use fewer receptive fields---27---for evaluation. We compare our results with SOTA under Protocol \#2. Table \ref{table:4} shows we achieved the best results in each action except the S3 of "Walking" due to corrupted mocap data. 
\subsubsection{Comparison with Temporal Convolutional  Networks}
\begin{figure}[t]
	\centering
	\includegraphics[width=0.48\textwidth]{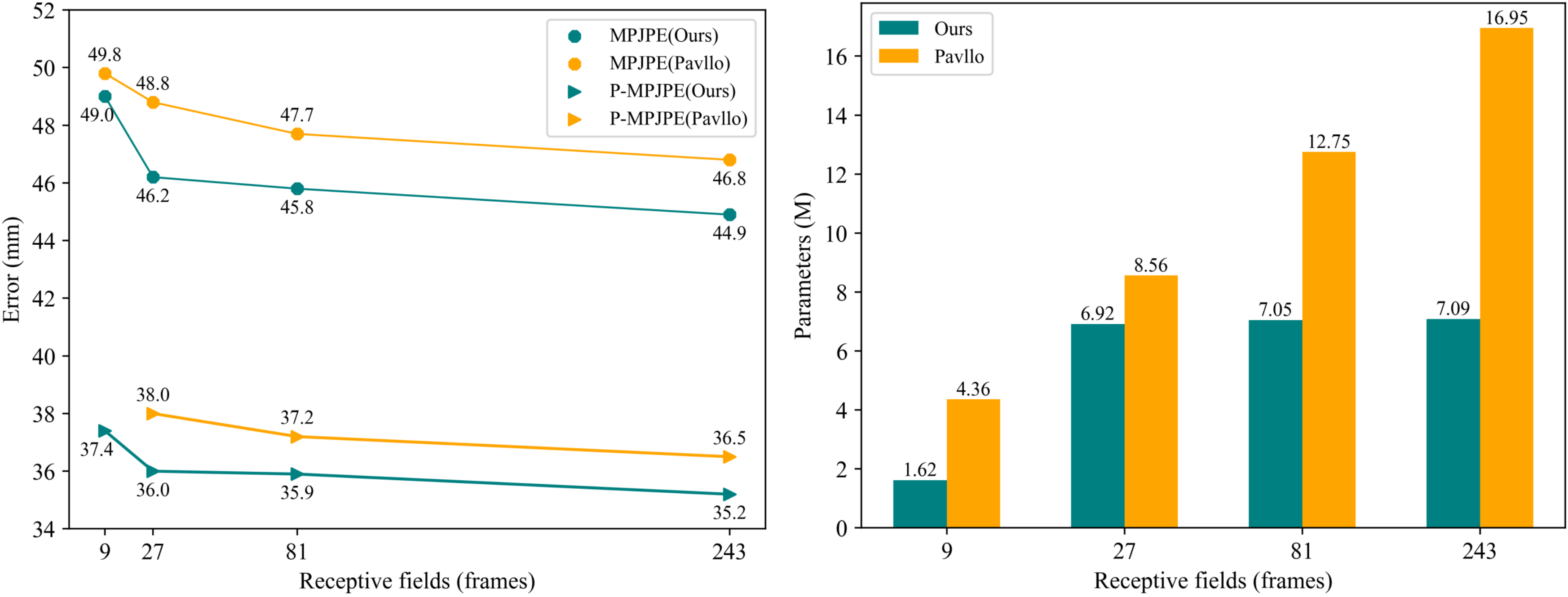}
	\caption{\textbf{Left}: Comparison with TCNs \cite{pavllo20193d} in different receptive fields on Human3.6M under protocol \#1 and \#2. \textbf{Right}: Comparison with the parameters of model.}
	\label{fig:comparision_tcns}
\end{figure}
Fig. \ref{fig:comparision_tcns} compares the number of parameters and the 3D pose estimation errors between TCNs \cite{pavllo20193d} and our various receptive field models. Like \cite{pavllo20193d}, we use CPNs as 2D pose detector for fair comparison. As can be seen on the left plot of Fig. \ref{fig:comparision_tcns}, we obtain smaller estimation errors for all receptive field combinations on Human3.6M under both protocols. Additionally, our model with 27 receptive fields is also slightly better than the TCNs with 243 receptive fields, which shows that the use of spatial information significantly contributes to reconstructing more accurate 3D poses. For the right side bar chart of Fig. \ref{fig:comparision_tcns}, we see that our model uses 62.9\%, 19.2\%, 44.7\%, and 58.2\% fewer parameters compared to the TCNs' 9, 27, 81, and 243 receptive field models respectively. This shows that interleaving of our spatial and temporal mechanisms results in a more efficient network for video pose estimation.
\subsection{Qualitative Results}
\begin{figure*}[thbp]
	\centering
	\includegraphics[width=0.90\textwidth]{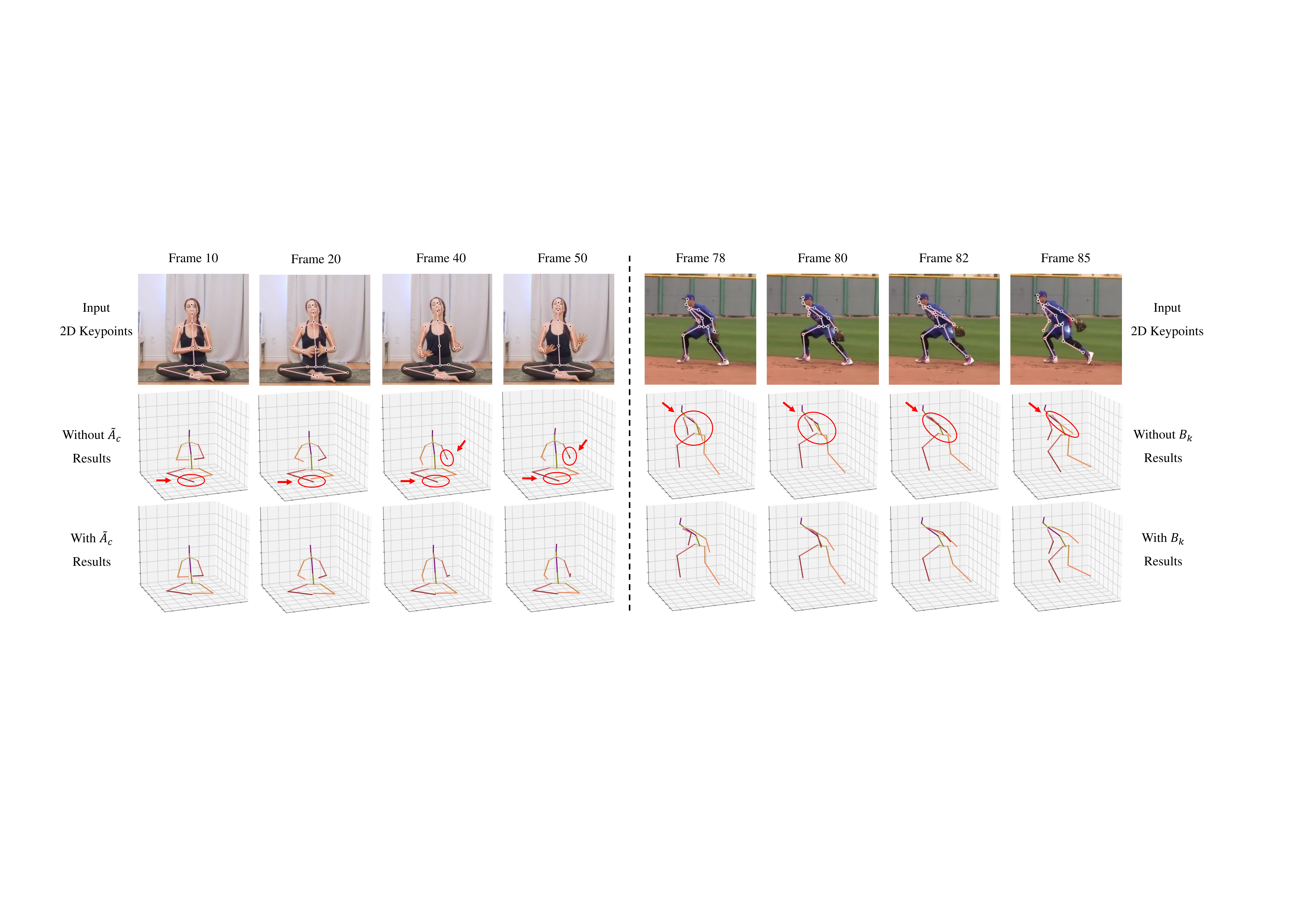}
	\caption{\textbf{Left}: Examples of results from our model with/without local kinematic connections $\widetilde{A}_{c}$. \textbf{Right}: Examples of results from our model with/without global adaptive matrix $B_{k}$. Wrong estimations are labeled in red circles.}
	\label{fig:local_global_att}
\end{figure*}
\begin{figure}[t]
	\centering
	\includegraphics[width=0.38\textwidth]{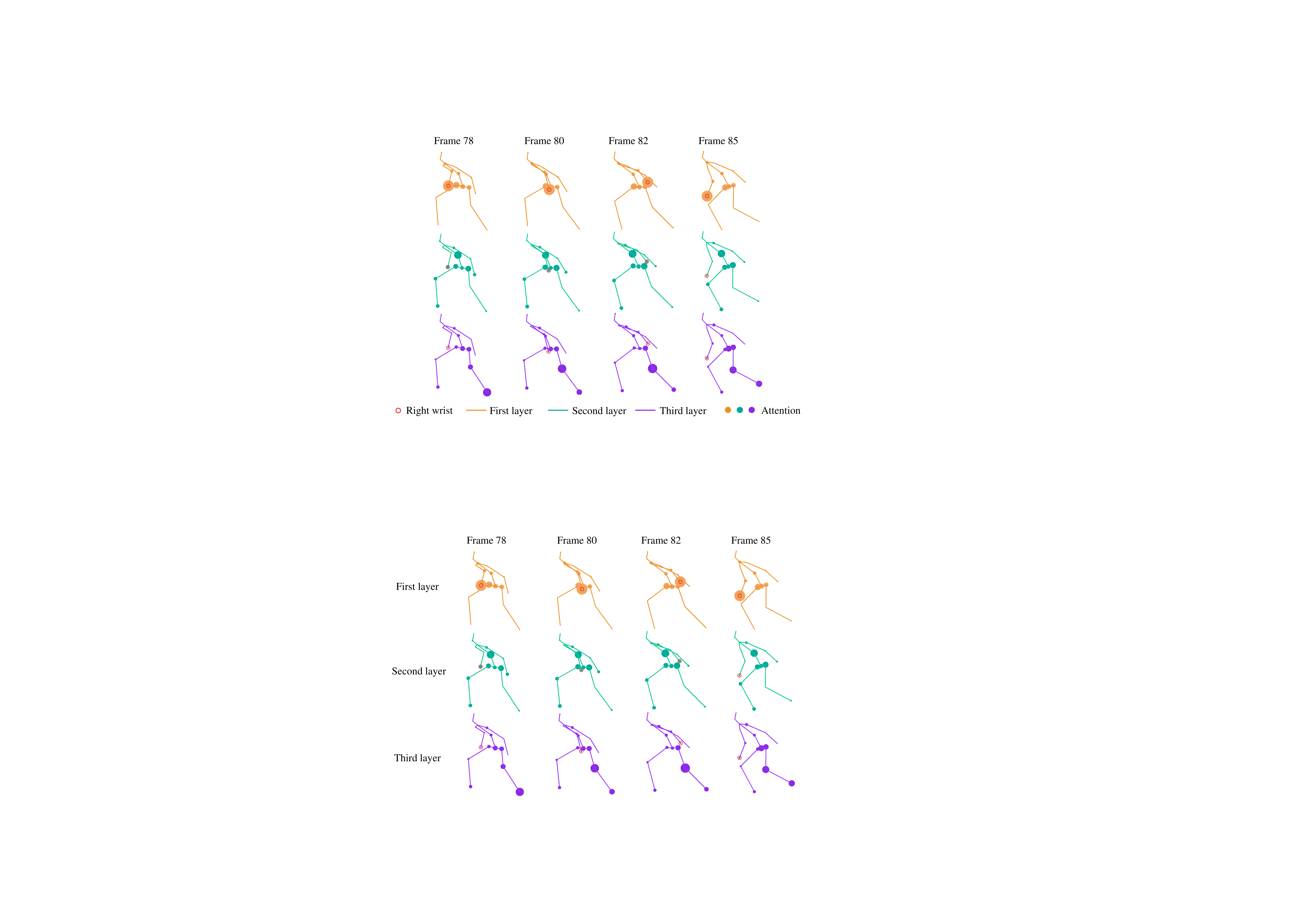}
	\caption{Visualization of global attention matrix $B_{k}$. The right wrist is set as the visualized joint. Three colored layers represent three different attention layers advancing from top to bottom. Circle mass indicates the relationship strength between the current and the visualized joint.}
	\label{fig:global_att_vis}
\end{figure}
\subsubsection{Significance of Spatial Semantics}
These studies do 3D pose estimation on YouTube videos using our 27 receptive field mode. The left of Fig. \ref{fig:local_global_att} shows pose reconstructions with/without local kinematic connections $\widetilde{A}_{c}$ in Yoga videos, while the right shows pose reconstructions with/without the global adaptive matrix $B_{k}$ in baseball videos. Pose errors are circled in red. Qualitative analysis for Yoga shows that when local kinematic connections are not considered there is considerable location ambiguity for distal joints. For the baseball videos, even when working with erroneous 2D input poses due to occlusions, the global posture semantics work effectively with temporal continuity to mitigate self-occlusion effects and yield accurate and smooth poses.
\subsubsection{Visualization of Global Attention Matrix $B_{k}$}
To further understand the construction of global semantics in occluded joints, we visualize our model's global attention matrices $B_{k}$ on the baseball case. Fig. \ref{fig:global_att_vis} is the visualization of the sample represented by skeleton graphs, where the circle mass indicates the strength of the relationship between the current joint and the right wrist in matrices. Skeleton graphs of each layer contain the average result across the multi-head attention of Eqtn. \ref{eqtn:globalGraph}. From the relationship visualized by the three-layer skeleton graphs, we see that the global graph attention tends to establish strong connections with the spine, left hip and right hip---joints close to the root joint. We speculate that the position information of these joints is easier to predict and stabilizes the reconstruction of the valid 3D structure. Apart from the aforementioned relationships, the skeleton graph in the first layer focuses on self-connections. For the 2nd and 3rd layers, strong relationships are constructed with the joints along the left and right leg. We argue that higher layers convey higher-level posture semantics that contribute to modeling the non-local spatial configuration constraints.
\subsubsection{Reconstruction of Special Case}
\begin{figure}[t]
	\centering
	\includegraphics[width=0.49\textwidth]{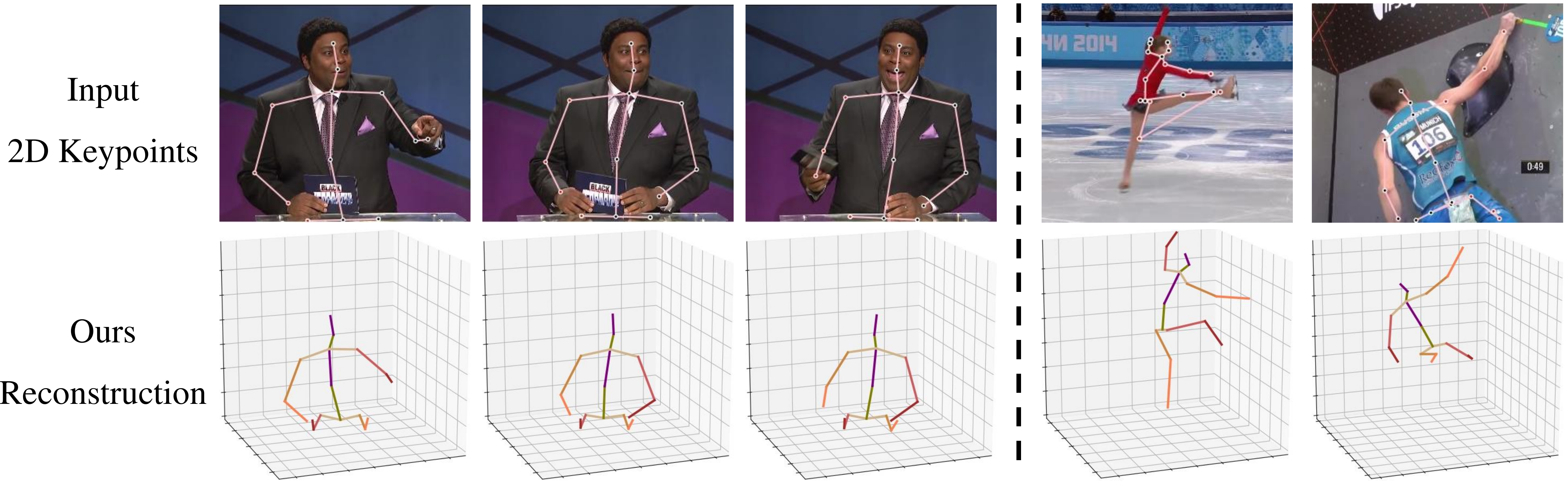}
	\caption{\textbf{Left}: Examples of results from our model reconstructed valid 3D structure from a half body. \textbf{Right}: Two failure cases caused by big 2D detection error and long-term heavy occlusion respectively.}
	\label{fig:half_body_failure_case}
\end{figure}
\begin{table}[t]
	\centering
	\caption{Test time of estimation speed}
	\label{table:5}
	\resizebox{0.48\textwidth}{!}{
	\begin{tabular}{l|ccc|ccc}
		\toprule[1pt]
		GAST-Net         & \multicolumn{3}{c|}{Layer-by-layer inference} & \multicolumn{3}{c}{Single-frame inference} \\ \midrule[0.5pt]
		Receptive fields & 27             & 81            & 243          & 27           & $ $ 81          & 243          \\ \midrule[0.5pt]
		Frames per second              & 1270           & 1120          & 960          & 74           & $ $ 56          & 45           \\ \bottomrule[1pt]
	\end{tabular}
	}
\end{table}
We also consider situations in which cameras only capture a person's upper body (common across applications, \eg human-robot interaction). We conducted experiments that reveal that GAST-Net yields reasonable 3D pose reconstructions as shown on the left of Fig. \ref{fig:half_body_failure_case}. Note that whilst GAST-Net is only trained on whole bodies, the model effectively reconstructs upper body test data it has not seen before. The right side of Fig \ref{fig:half_body_failure_case} shows two failure cases caused by large 2D detection errors (ice skating) as well as significant occlusions over long-term periods of time (wall climbing).
\subsection{Testing the Speed of 3D Pose Estimation}
We implemented our model with different inference modes and receptive fields to test the speed of 2D-to-3D video pose estimation. The speed time results are summarized in Table \ref{table:5}. These tests used an Intel CORE CPU@2.20GHz(6 cores) laptop and an NVIDIA GTX 1060 GPU. We use native Python without parallel optimization for inference. Since layer-by-layer inference enables parallel processing on the input frames, the estimation speed is faster compared to single-frame inference. To intuitively understand the estimation speed of single-frame inference from RGB videos, we adopt YOLOv3 (160$\times$160) \cite{redmon2018yolov3} and SORT \cite{Bewley2016_sort} for human detection and tracking, HRNet (256$\times$192) \cite{sun2019deep} for 2D pose estimation, and GAST-Net (27 receptive fields) for 2D-to-3D pose reconstruction. Experiments show that our top-down video pose estimation achieves 11 fps for a single person with the same test environment.

\section{Conclusion}
In this work, we presented a real-time 3D pose estimation methodology that simultaneously and flexibly captures varying spatio-temporal sequences. The proposed graph attention blocks, effectively model the symmetrical hierarchy of 2D skeleton as well as global postural constraints, are synergistically combined with temporal dependencies to better mitigate depth ambiguity and resolve self-occlusion. Qualitative results show that our approach also generalizes to 3D pose estimation in the half upper body, which helps to close-range interactive applications (\emph{e.g.}, human-robot interaction).
\bibliographystyle{IEEEtran}
\bibliography{references}
\end{document}